\documentclass[runningheads]{llncs}
\usepackage[T1]{fontenc}
\usepackage{bbding}
\usepackage{graphicx}
\begin{document}
\title{Traditional Chinese Medicine Case Analysis System for High-Level Semantic Abstraction: Optimized with Prompt and RAG}
\titlerunning{Optimizing TCM Case Analysis with Prompt Engineering and RAG}
\author{Peng Xu\inst{1} \and
Hongjin Wu\inst{2} \and
Jinle Wang \inst{3 (}\Envelope\inst{)} \and Rongjia Lin\inst{4} \and Liwei Tan\inst{5 (}\Envelope\inst{)}  }
\authorrunning{P. Xu et al.}
\institute{Chongqing Jiudun Technology Co., Ltd., Chongqing, China \and
School of Mechanical Science \& Engineering, Huazhong University of Science and Technology, Wuhan, China \and School of Nursing, Shanghai Jiao Tong University, Shanghai, China \email{wangjinlee123@163.com} \and Xinjiang Medical University, Ürümqi, China \and School of Mathematical Sciences, Shanghai Jiao Tong University, Shanghai, China
\email{TGS123@sjtu.edu.cn}}

\maketitle              % typeset the header of the contribution
\begin{abstract}
This paper details a technical plan for building a clinical case database for Traditional Chinese Medicine (TCM) using web scraping. Leveraging multiple platforms, including 360doc, we gathered over 5,000 TCM clinical cases, performed data cleaning, and structured the dataset with crucial fields such as patient details, pathogenesis, syndromes, and annotations. Using the $Baidu\_ERNIE\_Speed\_128K$ API, we removed redundant information and generated the final answers through the $DeepSeekv2$ API, outputting results in standard JSON format. We optimized data recall with RAG and rerank techniques during retrieval and developed a hybrid matching scheme. By combining two-stage retrieval method with keyword matching via Jieba, we significantly enhanced the accuracy of model outputs.
\end{abstract}
\section{Introduction}
Traditional Chinese Medicine (TCM) has been an integral part of Chinese culture for millennia, offering unique diagnostic and therapeutic approaches distinct from Western medicine. Central to TCM is syndrome differentiation, a complex process requiring practitioners to synthesize diverse clinical information to identify underlying pathologies and formulate appropriate treatments. This intricate reasoning process challenges modernization and integration with contemporary medical practices.

Recent developments in Artificial Intelligence (AI), particularly with large language models (LLMs), have shown promising applications in fields requiring complex language understanding and reasoning, including healthcare and Traditional Chinese Medicine (TCM). Studies have demonstrated that LLMs such as Zhongjing and Qibo, which are tailored for TCM, can process extensive datasets, uncover underlying patterns, and facilitate new insights into TCM diagnosis and treatment\cite{Yang2023}. These AI models can enhance diagnostic accuracy, create standardized treatment approaches, and support integrating traditional practices with modern technological advancements\cite{Yue2024,Ren2022}.

Several studies have explored the application of AI in TCM. For instance, Yang et al. (2023) introduced Zhongjing, a Chinese medical LLM enhanced through expert feedback and real-world multi-turn dialogues, demonstrating improved capabilities in TCM contexts\cite{Yang2023}. 
 Similarly, Zhang et al. (2024) developed Qibo, an LLM tailored for TCM, addressing challenges such as the divergence between TCM theory and modern medicine, and the scarcity of specialized corpora\cite{Zhang2024}.  These efforts underscore the potential of AI to simulate TCM's clinical reasoning processes, thereby elucidating its theoretical framework and facilitating its global dissemination.

Despite these advancements, challenges remain in developing AI models capable of performing syndrome differentiation with the depth and nuance of experienced TCM practitioners. Existing benchmarks, such as ShenNong-TCM-EB and Qibo-benchmark, primarily focus on basic TCM knowledge and lack the complexity required to evaluate diagnostic reasoning in syndrome differentiation\cite{Yue2024}.  This gap highlights the need for comprehensive, objective, and systematic evaluation datasets to assess and enhance AI models' capabilities in TCM syndrome differentiation.

Our study aims to construct a high-quality TCM clinical case database tailored to syndrome differentiation reasoning in this context. We collected over 5,000 clinical TCM cases from sources such as 360doc by leveraging web scraping and structured data cleaning techniques. We utilized Baidu's ERNIE Speed 128K API to remove irrelevant information and structure the data into fields, including patient details, syndrome pathology, diagnosis, and treatment notes. For retrieval and reasoning, we implemented hybrid matching schemes combining vector similarity-based matching using gte-Qwen2-1.5B-instruct and keyword matching via jieba segmentation\cite{Zhang2019}, then rerank the results from both retrievers by gte-passage-ranking-multilingual-base\cite{zhang2024mgte}, achieving improved accuracy and relevance in information retrieval\cite{Weisz2023,Webb2023}. Answer generation was accomplished using DeepSeekv2 API\cite{DeepSeekAI2024} with prompt optimization, outputting results in JSON format to ensure data standardization.

Our approach also involved testing Retrieval Augmented Generation (RAG) and rerank techniques, enhancing model responses by integrating initial retrieval results with reordering strategies prioritizing higher-quality outputs. Through these advancements, we constructed a high-quality TCM clinical case database that provides valuable resources for researchers and practitioners while setting a standardized evaluation benchmark for TCM language models.

The remainder of this paper is organized as follows: Section 2 reviews related work, highlighting the advancements and challenges in TCM data processing and retrieval. Section 3 outlines the methodology, including data preparation, prompt optimization, and the integration of RAG and Jieba segmentation. Section 4 presents the experimental results, followed by the conclusion in Section 5, which summarizes the findings and discusses future directions.
\section{Related Work}
\subsection{Construction of TCM Clinical Case Databases}

The construction of clinical case databases in Traditional Chinese Medicine (TCM) has become increasingly important in facilitating research and practice by standardizing and consolidating clinical information. The China TCM Clinical Case Database by the China Association of Chinese Medicine is one such repository that compiles case studies from renowned TCM experts, aiming to preserve and share TCM knowledge with a broader audience\cite{Ren2022,timmurphy.org2}. Similarly, studies like those on integrated TCM clinical data platforms focus on managing and analyzing vast amounts of clinical data, providing a structured resource for clinical and research purposes in TCM\cite{Zhang2022,Ren2022}.

Our approach builds upon these existing databases by expanding the scope of data sources and integrating advanced retrieval techniques to enhance the accessibility and usability of TCM knowledge\cite{Zhang2022a}. Unlike existing work primarily focusing on case compilation, our system incorporates robust data cleaning, text vectorization, and hybrid retrieval techniques to provide  a highly interactive and accurate retrieval experience for TCM practitioners and researchers\cite{Zhang2024,Yue2024}.

\subsection{Data Cleaning and Structuring for TCM Applications}

Data preprocessing, especially data cleaning and structuring, is critical for building high-quality TCM databases. The Chinese Medicine Real-World Data Collection Specification guides data collection methods, database construction, and integration, ensuring high-quality, standardized data across diverse TCM sources\cite{Li2022}. 
Leveraging such standards, our work utilizes the $Baidu\_ERNIE\_Speed\_128K$ API\cite{Sun2019} to harness the understanding capabilities of large language models. This enables us to perform two critical tasks: first, splitting concatenated low-quality TCM case data into individual cases; and second, filtering out irrelevant information and symbols while extracting essential information, thereby ensuring consistency and quality across the dataset.

Compared to existing efforts focusing on basic data cleaning, our approach also involves structuring TCM case data into essential categories, including patient background, pathogenesis, syndromes, and doctor’s notes. This structured format enables a more organized and efficient retrieval process, ensuring TCM researchers can access well-organized data for further analysis and study.

\subsection{Hybrid Retrieval Techniques for Enhanced Search Accuracy}

Traditional text-based retrieval methods often need to be revised when applied to complex, domain-specific datasets like TCM clinical cases. In response, hybrid retrieval techniques that combine semantic vector similarity with keyword-based matching have gained traction\cite{Sarmah2024}. Studies such as Tencent Cloud's hybrid search applications demonstrate the effectiveness of combining vector-based search with traditional keyword matching to improve search relevance and accuracy\cite{hybrid-search-is-a-method-to-optimize-rag-implementation-98d9d0911341}. Similarly, Milvus hybrid search tutorials highlight the benefits of integrating dense and sparse vector matching to enhance retrieval precision in specialized fields\cite{hybrid-search-with-milvus}.

Our work builds upon these advancements by utilizing $gte\_Qwen2-1.5B-instruct$\cite{Hui2024} for semantic vector generation and Jieba segmentation for keyword extraction, creating a hybrid retrieval system optimised explicitly for TCM data. By combining these methods, our system ensures semantic accuracy and keyword relevance, addressing the unique needs of TCM retrieval. Additionally, we employ retrieval-augmented generation (RAG) techniques\cite{Lewis2020} and reranking strategies\cite{Gao2024} to improve the overall recall and ranking quality, further refining the retrieval process for complex TCM queries.

\subsection{Prompt Engineering and Answer Generation in TCM}

In specialized domains like TCM, prompt engineering has become essential for guiding language models in complex reasoning tasks. Recent research shows that Chain-of-Thought (CoT) prompts\cite{Wei2022} can significantly improve reasoning accuracy in natural language models, especially in domains that require detailed knowledge like medicine\cite{Perry1990}. Building on this, we employ prompt optimization techniques with the DeepSeekv2 API to generate high-quality answers in JSON format, allowing structured responses for TCM queries. Our iterative prompt design process ensures that generated answers align closely with the diagnostic and interpretive standards expected in TCM.
\section{Method}
\subsection{Data prepare}
\subsubsection{Data Collection}
The data for this study were sourced from multiple reputable Traditional Chinese Medicine (TCM) knowledge-sharing platforms, including 360doc and other online databases. These platforms provided access to a diverse range of TCM clinical cases, ensuring a comprehensive dataset for analysis. In total, over 5,000 TCM clinical cases were collected to guarantee sufficient breadth and depth of information. The data fields encompass essential attributes such as patient demographics, pathogenesis, syndromes, and commentary, among other critical details, which lay the foundation for subsequent analysis and application.

\subsubsection{Data Cleaning and Standardization}  
To ensure high-quality data, a meticulous data cleaning process was undertaken using the $Baidu\_ERNIE\_Speed\_128K$ API. This tool was employed to eliminate redundant and irrelevant information, thereby enhancing the dataset's reliability and usability. Following the cleaning process, the data underwent a comprehensive standardization and structuring phase. This step involved normalizing the field definitions across different data sources to achieve consistency and readability. The result was a well-organized database that meets the requirements of standardized data formats, ensuring compatibility for advanced analyses and machine-learning applications.

\subsubsection{Data Integration and Accessibility}  
The processed and structured database is aligned with academic standards for TCM clinical case studies and serves as a resource for further research and development. The dataset is now hosted and made accessible on a dedicated platform for TCM clinical knowledge, available at https://cccl-tcm.cacm.org.cn/article?lang=zh. This online resource facilitates seamless access for researchers and practitioners interested in exploring diverse TCM clinical case studies, enabling deeper insights into evidence-based TCM practices.

\subsection{Chain of Thought}
Chain-of-Thought (CoT) prompting is an effective method to enhance reasoning capabilities in TCM syndrome differentiation by structuring the diagnostic process into clear steps. Starting from clinical information, CoT guides the inference of pathomechanisms and the determination of syndromes through systematic reasoning paths. For instance, given a patient’s symptoms such as retrosternal pain, nausea, and acid regurgitation, CoT enables the model to extract key clinical features, infer underlying mechanisms like "stomach disharmony" or "liver qi stagnation" and classify syndromes such as "liver-stomach disharmony" This approach ensures logical and accurate diagnostic outcomes, aligning with the structured thinking required in TCM and improving the model's ability to handle complex causal relationships in diagnosis.

\begin{figure}[htpt!]
\includegraphics[width=1.0\textwidth]{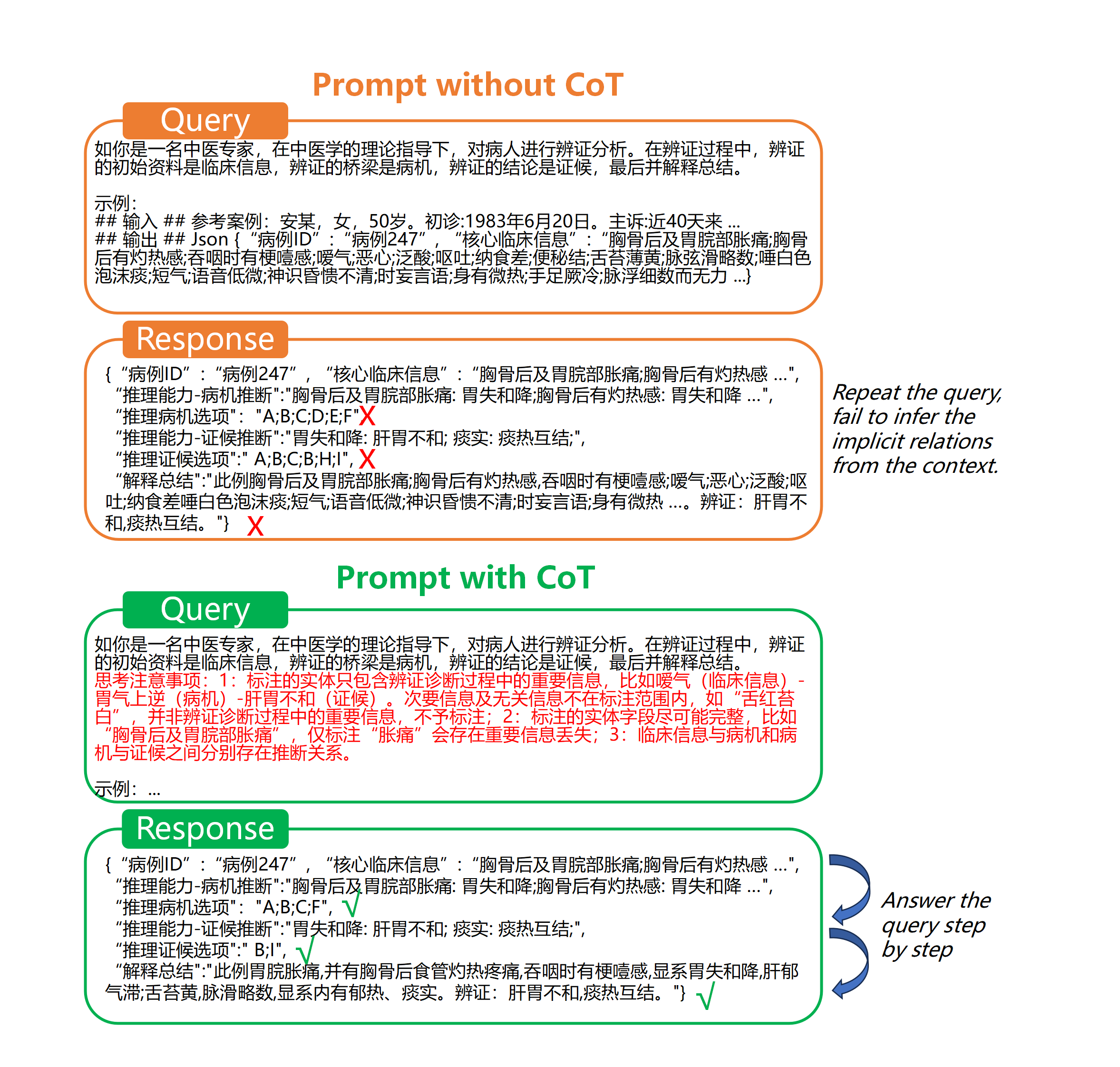}
\centering
\caption{An example of constructing chain-of-thought prompts in the TCM syndrome differentiation task involves integrating reasoning pathways into the diagnostic process. Compared with the original prompt, the chain-of-thought prompt adds the corresponding reasoning path for extracting clinical information, inferring pathomechanisms, and determining syndromes before outputting the final diagnosis, ensuring a systematic and logical diagnostic approach.} \label{fig1}
\end{figure}

Fig.\,\ref{fig1}. illustrates the difference between a prompt without Chain-of-Thought (CoT) reasoning and one with CoT in TCM syndrome differentiation tasks. In the "Prompt without CoT" the system repeats the query and fails to infer the implicit relations between clinical information, pathomechanisms, and syndromes, leading to incomplete or incorrect reasoning. In contrast, the "Prompt with CoT" demonstrates step-by-step reasoning, systematically extracting vital clinical features, inferring intermediate pathomechanisms, and deriving the appropriate syndrome classification. This approach enables logical and accurate outputs by explicitly guiding the reasoning process, showcasing how CoT prompting improves interpretability and performance in complex diagnostic tasks.

\subsection{Retrieval-Augmented Generation (RAG)}
\begin{figure}[htpt!]
\includegraphics[width=0.88\textwidth]{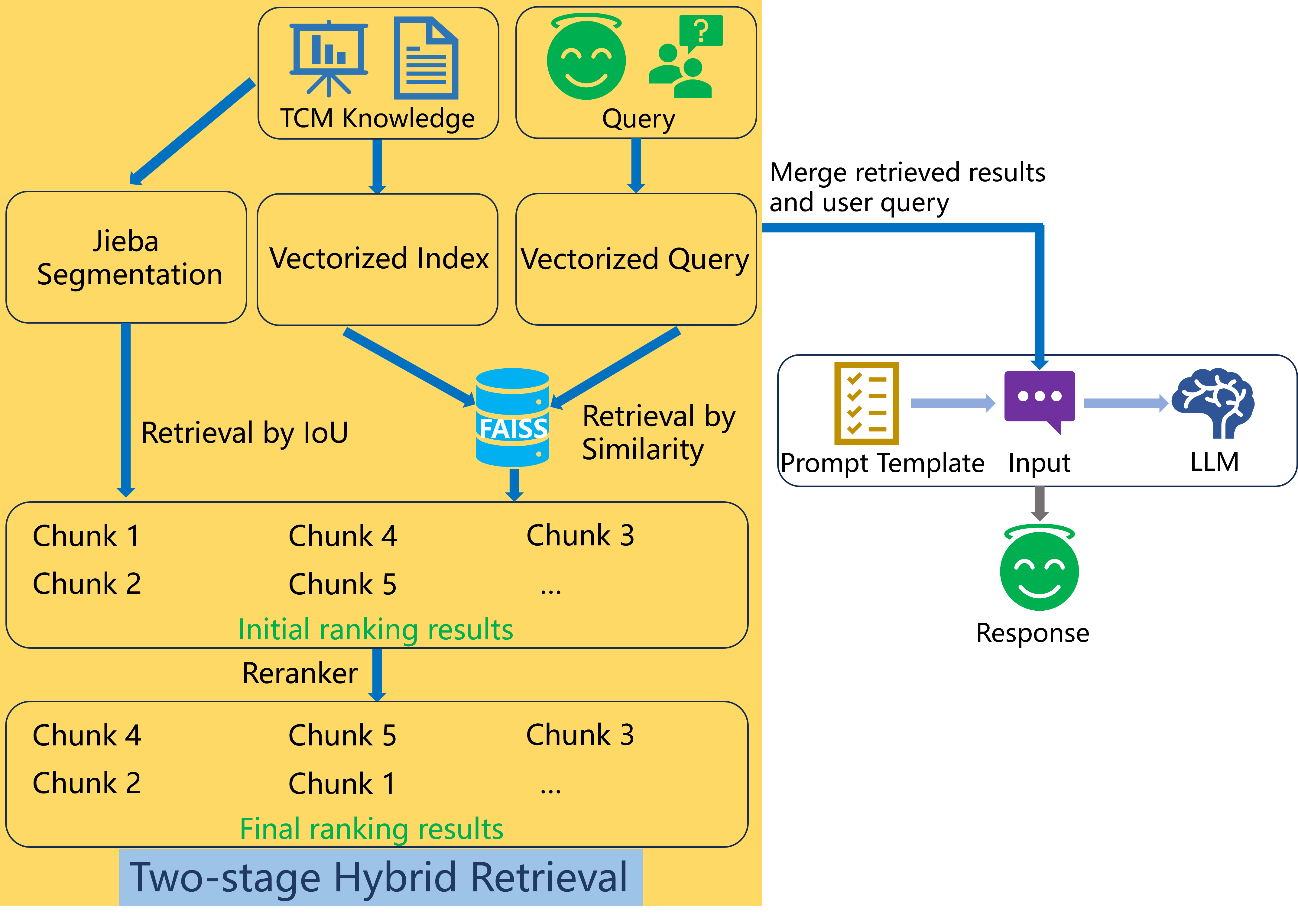}
\centering
\caption{Illustration of a two-stage retrieval process in a Retrieval-Augmented Generation (RAG) framework for TCM diagnosis, integrating user queries with external knowledge to enhance reasoning and output accuracy.} \label{fig2}
\end{figure}

Retrieval-augmented generation (RAG) offers a powerful approach to enhancing diagnostic reasoning in Traditional Chinese Medicine (TCM) by integrating external knowledge retrieval with advanced generative language models. This hybrid methodology addresses the limitations of standalone models by leveraging structured external data to augment the inference process, thereby improving the quality and reliability of diagnostic outputs. The workflow can be shown in Fig.\,\ref{fig2}.
\begin{itemize}
    \item \textbf{Two-Stage Retrieval Framework.}  The RAG framework employs a two-stage retrieval process to extract and rank relevant information from a pre-constructed TCM knowledge base. First, both the user query and the indexed knowledge are vectorized, with similarity-based ranking performed using tools such as FAISS (Facebook AI Similarity Search). The system retrieves a pool of knowledge chunks in the initial stage based on semantic similarity. Subsequently, these initial results are refined through a re-ranking mechanism to prioritize the most contextually relevant information for downstream tasks.
    
    \item \textbf{Integration with Generative Models.} The retrieved information is incorporated into the generative process by merging it with the user query through a structured prompt template. This enriched input is fed into a large language model (LLM), enabling it to generate contextually informed responses grounded in domain-specific knowledge.
    
    \item \textbf{Improved Diagnostic Reasoning.}  By incorporating external knowledge, RAG significantly enhances the model's ability to perform complex reasoning tasks, such as pathomechanism inference and syndrome differentiation in TCM. The retrieved information provides a robust foundation for reasoning, ensuring that the outputs—such as the inferred pathomechanism, syndrome classifications, and explanatory summaries—are accurate, interpretable, and aligned with TCM theoretical principles.
\end{itemize}
\subsection{Jieba segmentation}
In TCM diagnostic tasks, Jieba segmentation is crucial in preprocessing and analyzing textual data, particularly in developing retrieval-augmented generation (RAG) systems. Jieba is a widely used Chinese text segmentation tool that utilizes dictionary-based methods and Hidden Markov Models (HMM) for accurate word segmentation. It breaks down complex Chinese sentences into meaningful segments, allowing for effective keyword extraction and semantic analysis.

\begin{figure}[htpt!]
\includegraphics[width=0.88\textwidth]{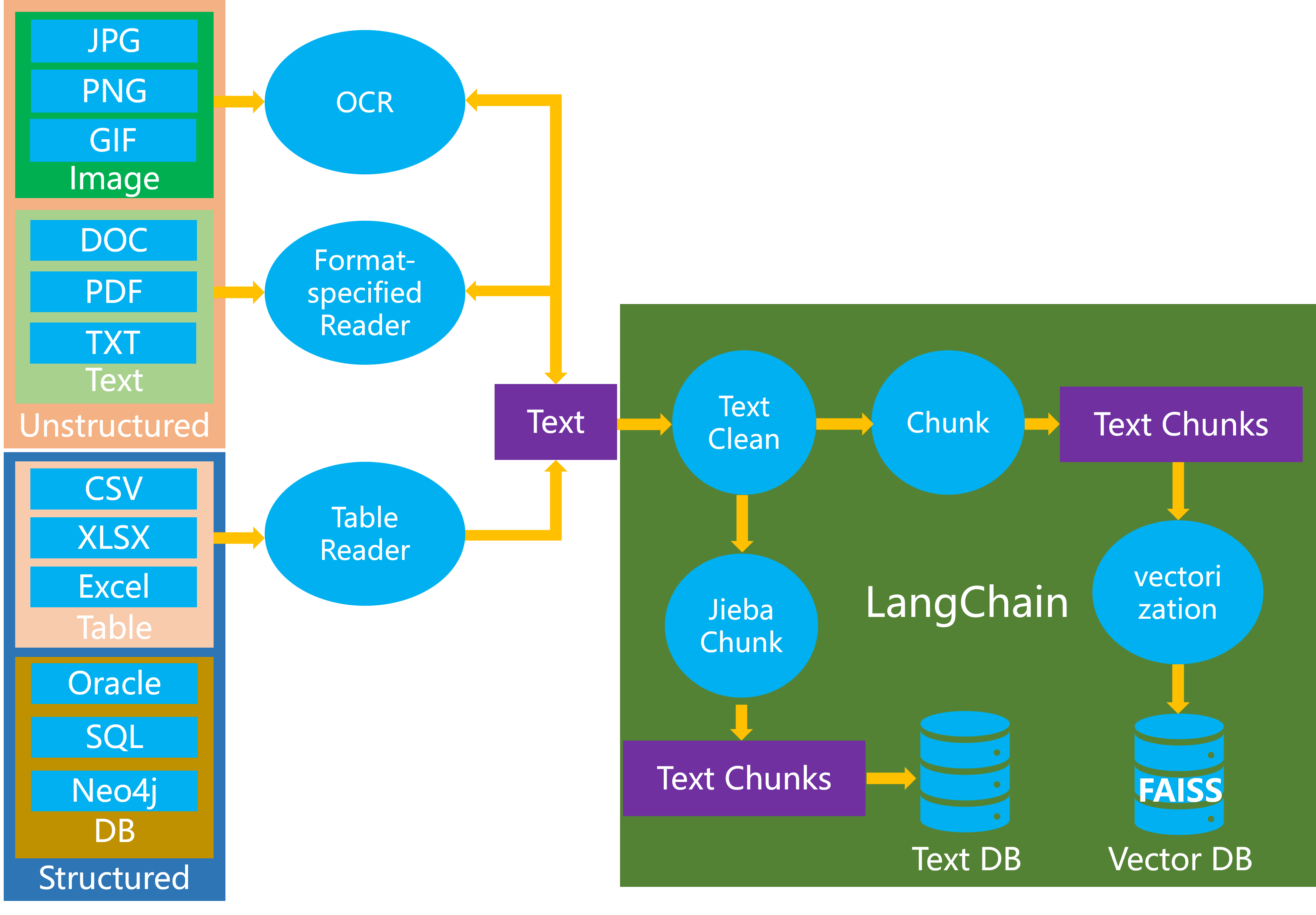}
\centering
\caption{Illustration of a preprocessing pipeline for structured and unstructured data, integrating OCR, text cleaning, jieba chunking, text segmentation, and vectorization for creating a searchable hybrid database.} \label{fig3}
\end{figure}

In the workflow illustrated in the Fig.\,\ref{fig3} and documentation, Jieba segmentation supports the construction of the TCM knowledge base by enabling precise extraction of key terms such as symptoms, pathomechanisms, and syndromes. By segmenting long and complex clinical texts into manageable entities, Jieba helps enhance the quality of the vectorized query and indexed knowledge. This segmentation, combined with RAG’s two-stage retrieval process, improves the relevance and accuracy of retrieved knowledge chunks.

Furthermore, Jieba segmentation enables mixed-matching strategies by integrating IoU-based matching with semantic similarity techniques, as highlighted in the provided materials. This approach ensures that both lexical and contextual nuances of TCM terms are captured, ultimately enhancing the model’s ability to perform accurate syndrome differentiation and pathomechanism inference. Through its robust segmentation capabilities, Jieba is a foundational tool in the preprocessing pipeline for TCM-specific RAG systems.
\subsection{Demonstration Ranking}
 The final step in constructing the prompt involves ranking selected demonstrations using a hybrid matching and reranking approach. In our method, we first process both the input query and the cases stored in the database using Jieba segmentation to extract meaningful keywords and semantic units. These are then matched with retrieval-augmented generation (RAG) techniques to identify semantically similar cases from the database. Once an initial ranking is generated, a reranking process is performed to refine the results based on contextual relevance and domain-specific importance. The highest-ranked case from the database, after reranking, is selected and incorporated into the prompt to provide a highly relevant example for the task. This approach ensures that the selected demonstration aligns closely with the input query and enhances the performance of downstream reasoning tasks.

\section{Results}
\subsection{Improving TCM Diagnosis with RAG}
\begin{figure}[htpt!]
\includegraphics[width=1.0\textwidth]{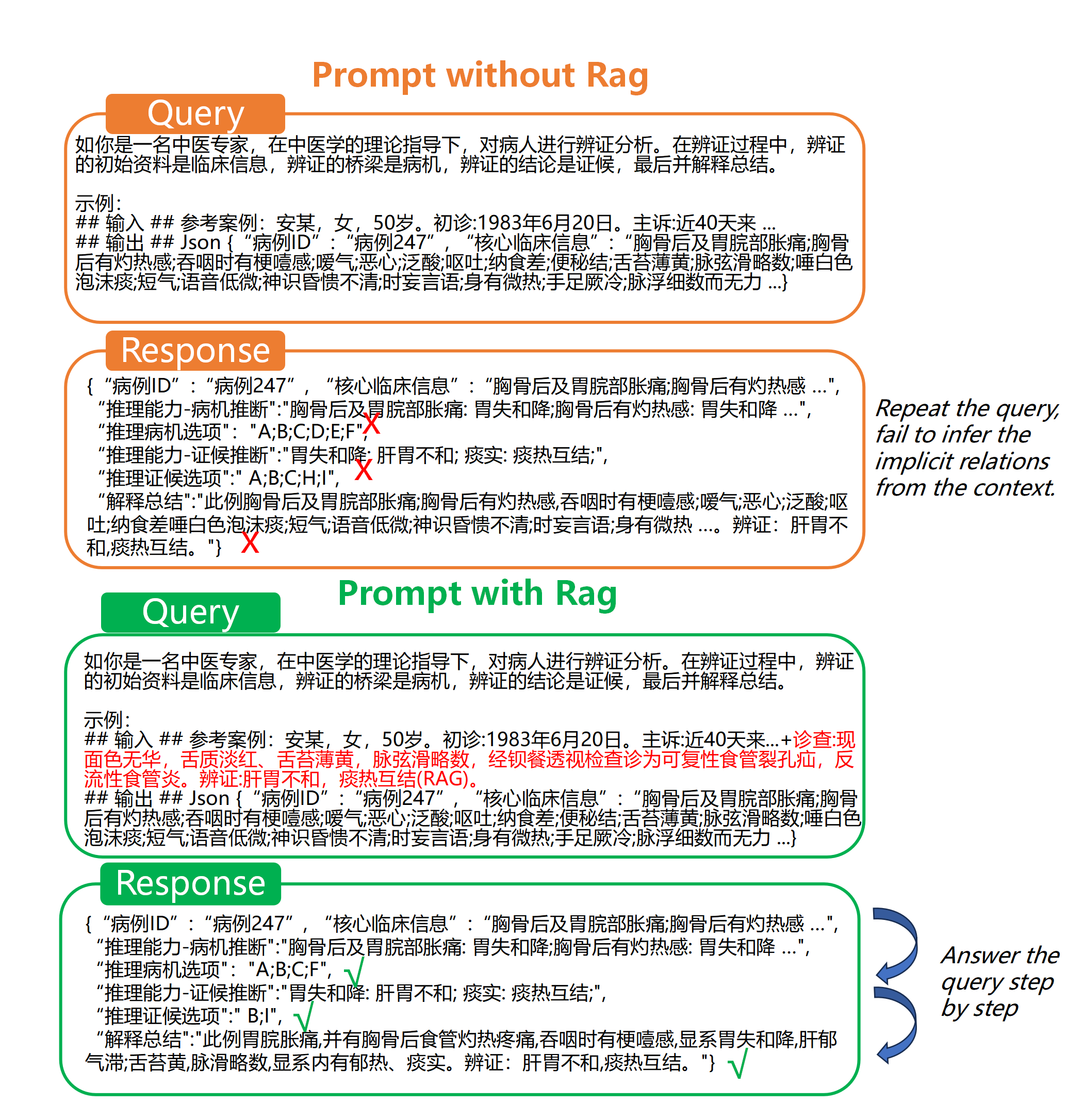}
\centering
\caption{Comparison of Diagnostic Reasoning with and without Retrieval-Augmented Generation (RAG) in Traditional Chinese Medicine Syndrome Differentiation.} \label{fig4}
\end{figure}

The Fig.\,\ref{fig4} illustrates the comparative effectiveness of two prompt engineering approaches-one without Retrieval-Augmented Generation (RAG) and the other incorporating RAG—in the context of Traditional Chinese Medicine (TCM) syndrome differentiation. The results demonstrate that including RAG significantly enhances the model's ability to extract, reason, and infer relationships within the diagnostic process.
In the "Prompt without RAG" approach, the model struggles to infer implicit relationships from the context and often repeats the input query without generating meaningful or logically coherent responses. For instance, in the example provided, the model incorrectly selects options for "pathogenesis reasoning" and provides a summary that lacks depth and alignment with TCM principles. This demonstrates the limitations of relying solely on the model's internal knowledge, as it fails to address the complexity of the task.
Conversely, the "Prompt with RAG" approach leverages external knowledge retrieval to supplement the model's reasoning process. By integrating relevant TCM case data and knowledge, the model can systematically extract clinical information, perform accurate pathogenesis and syndrome differentiation, and provide a well-structured, step-by-step explanation. For example, the model correctly identifies the relationships between clinical symptoms, Pathogenesis ("stomach disharmony and impaired descending" and "liver qi stagnation") and syndromes ("liver-stomach disharmony" and "phlegm-heat accumulation"), demonstrating improved reasoning accuracy and alignment with TCM diagnostic principles.
\subsection{Effect of RAG and Jieba segmentation}
The results in Tab.\ref{tab:rag_jieba_scores} demonstrate the comparative performance between naive overlap window and hybrid method with Jieba in handling traditional Chinese medicine (TCM) dataset tasks. While both methods show competitive results, Jieba slightly outperforms overlap window with a score of 37.05 compared to 36.15.
Jieba Segmentation: The higher score achieved by Jieba suggests its advantage in keyword-based matching tasks. Jieba's ability to perform efficient tokenization of Chinese text likely contributes to its strong performance in this context.
RAG (Retrieval-Augmented Generation): Although slightly behind Jieba, RAG performed well with a score of 36.15, showcasing its effectiveness in leveraging semantic retrieval. RAG's ability to retrieve contextually relevant information might make it more suitable for tasks requiring deeper semantic understanding.
The comparison highlights that while Jieba excels in tasks dependent on accurate tokenization and keyword matching, RAG's strength lies in semantic-driven retrieval. This suggests potential benefits in combining both methods to achieve optimal results, leveraging Jieba's precision in segmentation and RAG's contextual retrieval capabilities.
\begin{table}[h]
\centering
\caption{Comparison of Scores for Different Text Segmentation Methods}
\begin{tabular}{|l|c|}
\hline
\textbf{Method} & \textbf{Score} \\ \hline
Naive RAG            & 36.15          \\ \hline
Naive RAG + Jieba           & 37.05          \\ \hline
\end{tabular}
\label{tab:rag_jieba_scores}
\end{table}

\subsection{Performance comparison of LLMs in TCM clinical reasoning}
Based on the document provided, the experimental section emphasizes leveraging various AI models to process and analyze Chinese medicine clinical case data. The comparison of different base models highlights their respective performance across tasks, particularly focusing on classification and data generation accuracy.

\begin{table}[h]
\centering
\caption{Comparison of Scores for Different Models (Base Version Test)}
\begin{tabular}{|l|c|c|}
\hline
\textbf{Model}  & \textbf{Access}     & \textbf{Score} \\ \hline
GPT-4o          &   API    & 25.62          \\ \hline
DeepSeekV2      &   API    & 27.38          \\ \hline
Qwen2-72B       &   Local  & 25.32          \\ \hline
GLM-4           &   API    & 24.13          \\ \hline
Baidu-ERNIE-Speed-128K           &   API    & 26.16          \\ \hline
\end{tabular}
\label{tab:model_scores}
\end{table}
The experiments shown in Tab.\ref{tab:model_scores} evaluated the performance of several base models on specific tasks involving traditional Chinese medicine (TCM) clinical reasoning and dataset analysis. The models were tested using structured and cleaned TCM data from various sources, including over 5,000 case records. Advanced AI techniques, such as prompt optimization and mixed retrieval methods, were employed to enhance performance.
Among the tested models: DeepSeekV2 achieved the highest score of 27.38, reflecting its strong ability to handle structured TCM data and classification tasks effectively. ERNIE (26.16) and GPT-4o (25.62) also demonstrated competitive results, albeit slightly less effective than DeepSeekV2. Qwen2-72B scored 25.32, showing moderate efficiency in semantic understanding and classification. GLM-4, with a score of 24.13, lagged slightly behind the other models, indicating room for improvement in handling TCM-specific data.

The experimental results suggest that DeepSeekV2 is better suited for tasks requiring nuanced interpretation of TCM data, potentially due to its prompt engineering and fine-tuned parameters. Future work will focus on refining model inputs, enhancing retrieval accuracy, and expanding the dataset to improve generalizability and performance across models.

\section{Conclusion}
This study presents a comprehensive framework for enhancing Traditional Chinese Medicine (TCM) syndrome differentiation through advanced artificial intelligence techniques, focusing on integrating Retrieval-Augmented Generation (RAG) and Jieba segmentation. By leveraging over 5,000 structured clinical cases and optimizing prompts with the DeepSeekV2 API, the research has demonstrated significant improvements in the accuracy and relevance of diagnostic reasoning.
Key findings highlight that including RAG enables models to retrieve and utilize external domain-specific knowledge, thereby overcoming the limitations of standalone generative models. Combining vector-based semantic matching (gte\_Qwen2-1.5B-instruct), keyword extraction (Jieba segmentation) and reranker (gte-passage-ranking-multilingual-base), the two-stage hybrid retrieval approach ensures precise and contextually rich outputs. Moreover, the step-by-step reasoning enabled by Chain-of-Thought (CoT) prompts further aligns diagnostic outputs with TCM principles, enhancing interpretability and reliability.
Comparative experiments reveal that RAG significantly enhances reasoning tasks, improving the inference of pathogenesis and syndrome classification accuracy. For instance, the RAG-based approach successfully identifies critical relationships, such as linking "stomach disharmony" and "liver qi stagnation" with "liver-stomach disharmony" and "phlegm-heat accumulation." Additionally, performance comparisons between segmentation methods indicate that Jieba excels in token-based tasks, while RAG provides a deeper semantic understanding, suggesting the complementary nature of these techniques.
This research establishes a robust methodological foundation for modernizing TCM diagnosis through AI. Developing a standardized TCM case database and retrieval-enhanced reasoning models contributes to bridging traditional diagnostic practices with contemporary computational methodologies. Future work should explore expanding the dataset, integrating multi-modal data (e.g., imaging and lab reports), and refining retrieval mechanisms further to enhance the generalizability and efficacy of AI-assisted TCM diagnostics. This progress benefits the scientific understanding of TCM and fosters its global dissemination and integration into modern healthcare systems.
\bibliographystyle{splncs04}
\bibliography{ref}

\begin{thebibliography}{10}
\providecommand{\url}[1]{\texttt{#1}}
\providecommand{\urlprefix}{URL }
\providecommand{\doi}[1]{https://doi.org/#1}

\bibitem{timmurphy.org2}
Line spacing in latex documents. [Online], \url{https://www.popsci.com/article/technology/how-will-drones-change-sports/}

\bibitem{hybrid-search-with-milvus}
Batifol, S.: Getting started with hybrid search with milvus. [Online], \url{https://zilliz.com/blog/hybrid-search-with-milvus}

\bibitem{hybrid-search-is-a-method-to-optimize-rag-implementation-98d9d0911341}
Csakash: Hybrid search a method to optimize rag implementation. [Online], \url{https://medium.com/@csakash03/hybrid-search-is-a-method-to-optimize-rag-implementation-98d9d0911341}

\bibitem{DeepSeekAI2024}
{DeepSeek-AI}, {Liu}, A., {Feng}, B., {Wang}, B., {Wang}, B., {Liu}, B., {Zhao}, C., {Dengr}, C., {Ruan}, C., {Dai}, D., {Guo}, D., {Yang}, D., {Chen}, D., {Ji}, D., {Li}, E., {Lin}, F., {Luo}, F., {Hao}, G., {Chen}, G., {Li}, G., {Zhang}, H., {Xu}, H., {Yang}, H., {Zhang}, H., {Ding}, H., {Xin}, H., {Gao}, H., {Li}, H., {Qu}, H., {Cai}, J.L., {Liang}, J., {Guo}, J., {Ni}, J., {Li}, J., {Chen}, J., {Yuan}, J., {Qiu}, J., {Song}, J., {Dong}, K., {Gao}, K., {Guan}, K., {Wang}, L., {Zhang}, L., {Xu}, L., {Xia}, L., {Zhao}, L., {Zhang}, L., {Li}, M., {Wang}, M., {Zhang}, M., {Zhang}, M., {Tang}, M., {Li}, M., {Tian}, N., {Huang}, P., {Wang}, P., {Zhang}, P., {Zhu}, Q., {Chen}, Q., {Du}, Q., {Chen}, R.J., {Jin}, R.L., {Ge}, R., {Pan}, R., {Xu}, R., {Chen}, R., {Li}, S.S., {Lu}, S., {Zhou}, S., {Chen}, S., {Wu}, S., {Ye}, S., {Ma}, S., {Wang}, S., {Zhou}, S., {Yu}, S., {Zhou}, S., {Zheng}, S., {Wang}, T., {Pei}, T., {Yuan}, T., {Sun}, T., {Xiao}, W.L., {Zeng}, W., {An}, W., {Liu}, W., {Liang}, W., {Gao}, W.,
  {Zhang}, W., {Li}, X.Q., {Jin}, X., {Wang}, X., {Bi}, X., {Liu}, X., {Wang}, X., {Shen}, X., {Chen}, X., {Chen}, X., {Nie}, X., {Sun}, X., {Wang}, X., {Liu}, X., {Xie}, X., {Yu}, X., {Song}, X., {Zhou}, X., {Yang}, X., {Lu}, X., {Su}, X., {Wu}, Y., {Li}, Y.K., {Wei}, Y.X., {Zhu}, Y.X., {Xu}, Y., {Huang}, Y., {Li}, Y., {Zhao}, Y., {Sun}, Y., {Li}, Y., {Wang}, Y., {Zheng}, Y., {Zhang}, Y., {Xiong}, Y., {Zhao}, Y., {He}, Y., {Tang}, Y., {Piao}, Y., {Dong}, Y., {Tan}, Y., {Liu}, Y., {Wang}, Y., {Guo}, Y., {Zhu}, Y., {Wang}, Y., {Zou}, Y., {Zha}, Y., {Ma}, Y., {Yan}, Y., {You}, Y., {Liu}, Y., {Ren}, Z.Z., {Ren}, Z., {Sha}, Z., {Fu}, Z., {Huang}, Z., {Zhang}, Z., {Xie}, Z., {Hao}, Z., {Shao}, Z., {Wen}, Z., {Xu}, Z., {Zhang}, Z., {Li}, Z., {Wang}, Z., {Gu}, Z., {Li}, Z., {Xie}, Z.: {DeepSeek-V2: A Strong, Economical, and Efficient Mixture-of-Experts Language Model}. arXiv e-prints arXiv:2405.04434 (May 2024). \doi{10.48550/arXiv.2405.04434}

\bibitem{Gao2024}
{Gao}, J., {Chen}, B., {Zhao}, X., {Liu}, W., {Li}, X., {Wang}, Y., {Zhang}, Z., {Wang}, W., {Ye}, Y., {Lin}, S., {Guo}, H., {Tang}, R.: {LLM-enhanced Reranking in Recommender Systems}. arXiv e-prints arXiv:2406.12433 (Jun 2024). \doi{10.48550/arXiv.2406.12433}

\bibitem{Hui2024}
{Hui}, B., {Yang}, J., {Cui}, Z., {Yang}, J., {Liu}, D., {Zhang}, L., {Liu}, T., {Zhang}, J., {Yu}, B., {Dang}, K., {Yang}, A., {Men}, R., {Huang}, F., {Ren}, X., {Ren}, X., {Zhou}, J., {Lin}, J.: {Qwen2.5-Coder Technical Report}. arXiv e-prints arXiv:2409.12186 (Sep 2024). \doi{10.48550/arXiv.2409.12186}

\bibitem{Lewis2020}
{Lewis}, P., {Perez}, E., {Piktus}, A., {Petroni}, F., {Karpukhin}, V., {Goyal}, N., {K{\"u}ttler}, H., {Lewis}, M., {Yih}, W.t., {Rockt{\"a}schel}, T., {Riedel}, S., {Kiela}, D.: {Retrieval-Augmented Generation for Knowledge-Intensive NLP Tasks}. arXiv e-prints arXiv:2005.11401 (May 2020). \doi{10.48550/arXiv.2005.11401}

\bibitem{Li2022}
Li, X., Ren, J., Zhang, W., Zhang, Z., Yu, J., Wu, J., Sun, H., Zhou, S., Yan, K., Yan, X., Wang, W.: Ltm-tcm: A comprehensive database for the linking of traditional chinese medicine with modern medicine at molecular and phenotypic levels. Pharmacological Research  \textbf{178},  106185 (2022). \doi{https://doi.org/10.1016/j.phrs.2022.106185}, \url{https://www.sciencedirect.com/science/article/pii/S104366182200130X}

\bibitem{Perry1990}
Perry, C.A.: Knowledge bases in medicine: a review. Bulletin of the Medical Library Association  \textbf{78},  271--82 (Jul 1990)

\bibitem{Ren2022}
{Ren}, M., {Huang}, H., {Zhou}, Y., {Cao}, Q., {Bu}, Y., {Gao}, Y.: {TCM-SD: A Benchmark for Probing Syndrome Differentiation via Natural Language Processing}. arXiv e-prints arXiv:2203.10839 (Mar 2022). \doi{10.48550/arXiv.2203.10839}

\bibitem{Sarmah2024}
{Sarmah}, B., {Hall}, B., {Rao}, R., {Patel}, S., {Pasquali}, S., {Mehta}, D.: {HybridRAG: Integrating Knowledge Graphs and Vector Retrieval Augmented Generation for Efficient Information Extraction}. arXiv e-prints arXiv:2408.04948 (Aug 2024). \doi{10.48550/arXiv.2408.04948}

\bibitem{Sun2019}
{Sun}, Y., {Wang}, S., {Li}, Y., {Feng}, S., {Chen}, X., {Zhang}, H., {Tian}, X., {Zhu}, D., {Tian}, H., {Wu}, H.: {ERNIE: Enhanced Representation through Knowledge Integration}. arXiv e-prints arXiv:1904.09223 (Apr 2019). \doi{10.48550/arXiv.1904.09223}

\bibitem{Webb2023}
Webb, N.A., Bot, C., Charpinet, S., Contini, T., Jouve, L., Meheut, H., Mei, S., Mosser, B., Soucail, G.: {Gender and Precarity in Astronomy}. In: {Journ\'ees 2022 de la Soci\'et\'e Fran\c{c}aise d\textquoteright{}Astronomie \& d\textquoteright{}Astrophysique} (3 2023)

\bibitem{Wei2022}
{Wei}, J., {Wang}, X., {Schuurmans}, D., {Bosma}, M., {Ichter}, B., {Xia}, F., {Chi}, E., {Le}, Q., {Zhou}, D.: {Chain-of-Thought Prompting Elicits Reasoning in Large Language Models}. arXiv e-prints arXiv:2201.11903 (Jan 2022). \doi{10.48550/arXiv.2201.11903}

\bibitem{Weisz2023}
{Weisz}, J.D., {Muller}, M., {He}, J., {Houde}, S.: {Toward General Design Principles for Generative AI Applications}. arXiv e-prints arXiv:2301.05578 (Jan 2023). \doi{10.48550/arXiv.2301.05578}

\bibitem{Yang2023}
{Yang}, S., {Zhao}, H., {Zhu}, S., {Zhou}, G., {Xu}, H., {Jia}, Y., {Zan}, H.: {Zhongjing: Enhancing the Chinese Medical Capabilities of Large Language Model through Expert Feedback and Real-world Multi-turn Dialogue}. arXiv e-prints arXiv:2308.03549 (Aug 2023). \doi{10.48550/arXiv.2308.03549}

\bibitem{Yue2024}
{Yue}, W., {Wang}, X., {Zhu}, W., {Guan}, M., {Zheng}, H., {Wang}, P., {Sun}, C., {Ma}, X.: {TCMBench: A Comprehensive Benchmark for Evaluating Large Language Models in Traditional Chinese Medicine}. arXiv e-prints arXiv:2406.01126 (Jun 2024). \doi{10.48550/arXiv.2406.01126}

\bibitem{Zhang2024}
{Zhang}, H., {Wang}, X., {Meng}, Z., {Chen}, Z., {Zhuang}, P., {Jia}, Y., {Xu}, D., {Guo}, W.: {Qibo: A Large Language Model for Traditional Chinese Medicine}. arXiv e-prints arXiv:2403.16056 (Mar 2024). \doi{10.48550/arXiv.2403.16056}

\bibitem{Zhang2022}
Zhang, P., Shen, S., Deng, W., Mao, S., Wang, Y.: The construction model of the tcm clinical knowledge coding database based on knowledge organization. BioMed Research International  \textbf{2022}, ~1--7 (01 2022). \doi{10.1155/2022/2503779}

\bibitem{Zhang2022a}
Zhang, T., Huang, Z., Wang, Y., Wen, C., Peng, Y., Ye, Y.: Information extraction from the text data on traditional chinese medicine: A review on tasks, challenges, and methods from 2010 to 2021. Evidence-based complementary and alternative medicine : eCAM  \textbf{2022},  1679589 (2022)

\bibitem{Zhang2019}
Zhang, X., Wu, P., Cai, J., Wang, K.: A contrastive study of chinese text segmentation tools in marketing notification texts. Journal of Physics: Conference Series  \textbf{1302},  022010 (08 2019). \doi{10.1088/1742-6596/1302/2/022010}

\bibitem{zhang2024mgte}
Zhang, X., Zhang, Y., Long, D., Xie, W., Dai, Z., Tang, J., Lin, H., Yang, B., Xie, P., Huang, F., et~al.: mgte: Generalized long-context text representation and reranking models for multilingual text retrieval. arXiv preprint arXiv:2407.19669  (2024)

\end{thebibliography}

\end{document}